\documentclass[letterpaper, 10 pt, conference]{ieeeconf}  % Comment this line out if you need a4paper

\IEEEoverridecommandlockouts                              % This command is only needed if 
\usepackage{graphics} % for pdf, bitmapped graphics files
\usepackage{amsmath} % assumes amsmath package installed
\usepackage[T1]{fontenc}
\usepackage{lmodern}

\usepackage{ulem}
\usepackage{xcolor}
\usepackage{subcaption}
\usepackage[OT1]{fontenc} 
\title{\LARGE \bf
BLaVe-CoT: Consistency-Aware Visual Question Answering for Blind and Low Vision Users
}

\author{
Wanyin Cheng$^{1}$ and Zanxi Ruan$^{2*}$% <-this % stops a space
\thanks{*This work was not supported by any organization.}% <-this % stops a space
\thanks{$^{1}$Wanyin Cheng is with the School of Cyber Science and Engineering, Qufu Normal University, Qufu, China. {\tt\small accecwan@163.com}}%
\thanks{$^{2}$Zanxi Ruan is with the Department of Computer Science, University of Verona, Verona, Italy. {\tt\small zanxi.ruan@univr.it}}%
\thanks{$^{*}$Corresponding author: Zanxi Ruan.}%
}

\begin{document}

\maketitle
\thispagestyle{empty}
\pagestyle{empty}

%%%%%%%%%%%%%%%%%%%%%%%%%%%%%%%%%%%%%%%%%%%%%%%%%%%%%%%%%%%%%%%%%%%%%%%%%%%%%%%%
\begin{abstract}
Visual Question Answering (VQA) holds great potential for assisting Blind and Low Vision (BLV) users, yet real-world usage remains challenging. Due to visual impairments, BLV users often take blurry or poorly framed photos and face difficulty in articulating specific questions about what they cannot fully see. As a result, their visual questions are frequently ambiguous, and different users may interpret them in diverse ways. This leads to multiple valid answers, each grounded in different image regions—posing a mismatch with conventional VQA systems that assume a single answer and region. To bridge this gap, we present \textbf{BLaVe-CoT}, a VQA framework designed to reason about answer consistency in the face of ambiguity. Our method proposes diverse candidate answers using a LoRA-tuned BLIP-2 model, then grounds each answer spatially using PolyFormer, and finally applies a chain-of-thought reasoning module to assess whether the answers refer to the same or different regions. Evaluated on the VQA-AnswerTherapy benchmark, BLaVe-CoT outperforms previous methods and proves more robust to the ambiguity and visual noise common in assistive settings. This work highlights the need for VQA systems that can adapt to real human uncertainty and provide inclusive support for BLV users. To foster further research and accessibility applications, we have made the code publicly available at \href{https://github.com/Accecwan/BLaVe-CoT}{https://github.com/Accecwan/BLaVe-CoT}.
\end{abstract}


%%%%%%%%%%%%%%%%%%%%%%%%%%%%%%%%%%%%%%%%%%%%%%%%%%%%%%%%%%%%%%%%%%%%%%%%%%%%%%%%
\section{INTRODUCTION} % 4.25最后编辑
% 1. 引入场景：强调 BLV 群体的需求与挑战，这样会显得整体工作更有价值和意义
Visual Question Answering (VQA) is a challenging multi-modal task that requires understanding an image and answering a natural language question about it \cite{kuang2025natural}. Among the many real-world applications of VQA, one of the most impactful lies in assisting blind and low vision (BLV) users in interpreting their visual surroundings \cite{de2024vqask}. Recent services like Be My Eyes \cite{avila2016remote} demonstrate the high demand for such assistance. However, BLV-captured images often suffer from poor quality (blur, occlusion, unusual framing) and contain unfamiliar visual content (e.g., guide canes, tactile labels), making standard VQA systems underperform \cite{gonzalez2025towards}. More importantly, ambiguous images and unclear questions often admit several equally plausible answers, and concealing this diversity can mislead BLV users. Diverse user intents mean that one question may map to different image regions, resulting a mismatch with current VQA models that presume a single answer and a single grounding.

% Visual Question Answering (VQA)[cite] is an interdisciplinary task that combines image understanding and natural language processing, aiming to enable a model to answer natural language questions based on image content. In practice, the answers to these questions are typically provided by a large group of crowd workers. One of the core challenges in VQA lies in the fact that different individuals, due to subjective differences in perceiving and interpreting the image and question, may provide different answers to the same visual question. Previous studies [] have shown that answer variability is common in VQA datasets. For instance, in several well-known VQA datasets, more than half of the visual questions lead to diverse answers. This phenomenon is not coincidental but results from a combination of factors, including the complexity of the image itself, the ambiguity in the question's semantics, individual cognitive biases, the openness of the task, and even the approximation of word meanings. These factors together contribute to the objective reality that "one question can have multiple reasonable answers" in VQA systems.
% 视觉问答（Visual Question Answering, VQA）[]是一项结合图像理解与自然语言处理的综合性任务，其目标是让模型能够基于图像内容回答与之相关的自然语言问题。在实际的数据构建过程中，答案通常由大量众包标注者提供。然而，VQA面临的一个核心挑战在于，由于人类在感知、理解图像及解读问题时存在主观差异，即使面对相同的视觉输入与提问内容，不同的标注者也可能给出多种不同的答案。已有研究表明[]，答案不一致的情况在VQA数据集中广泛存在。例如，在多个主流VQA数据集中，超过一半的问题会导致多样化的答案[]。这一现象的产生并非偶然，而是受到多种因素[]的综合影响：包括图像本身的复杂性、问题语义的模糊性、个体认知偏好、任务的开放性，甚至词义的近似性等。这些因素共同导致了VQA系统中“一个问题可能对应多个合理答案”的客观现实[]。
\begin{figure}
    \centering
    \includegraphics[width=0.8\linewidth]{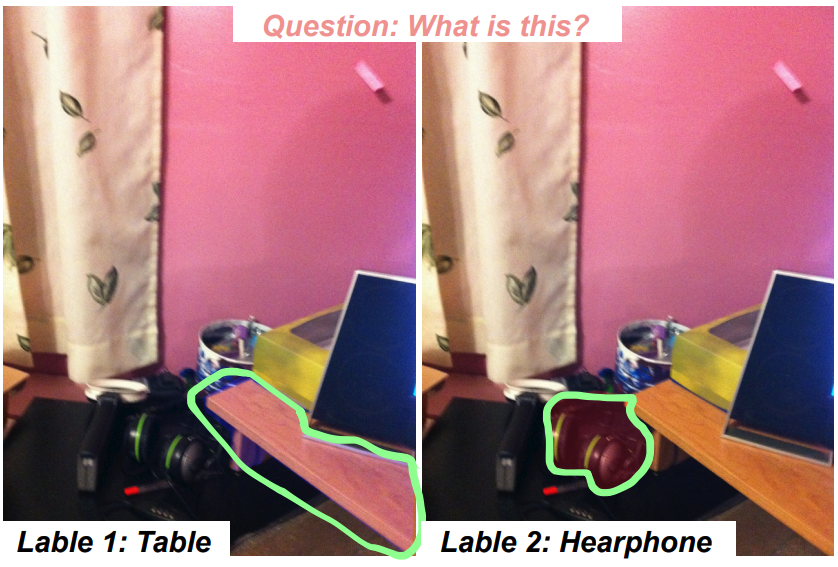}
\caption{
\textbf{Motivating example from a BLV VQA scenario.} Due to perceptual and motor limitations, BLV users often capture poorly framed images and pose underspecified questions.
}
    \label{fig:teaser}
\end{figure}
% 2. 引出 VQA-AnswerTherapy：强调其创新性，因为它是一个盲人视角的数据集
To better study the challenges of answer diversity in VQA for BLV users, researchers introduced the VQA-AnswerTherapy dataset \cite{chen2023vqa}. It provides visual groundings for multiple valid answers per question, enabling fine-grained analysis of answer variability. Most images come from the real-world VizWiz dataset \cite{gurari2018vizwiz}, reflecting common assistive scenarios with blurry, poorly framed, and subjective visual content. VQA-AnswerTherapy shows that many BLV questions trigger distinct answer regions (multi-grounding), not just paraphrases over one region (single-grounding). For example, when asked \textit{"What does this say?"}, different users may focus on different parts of a product package, such as brand names, slogans, or ingredients, resulting in multiple reasonable answers grounded in distinct regions. To further illustrate this challenge, Figure~\ref{fig:teaser} presents a typical case from our dataset. Faced with the vague question \textit{``What is this?''}, the annotators give two valid answers: ``Table'' and ``Headphone'', each grounded in a different region. This discrepancy is not accidental: due to limited visual access, BLV users often misframe images and struggle to specify spatial intent in their questions. Such ambiguity exposes a core limitation of conventional VQA models and underscores the need for a framework that reasons over visual-semantic consistency rather than relying on a fixed ground truth.

% In order to differentiate the inherent challenges posed by the visual question itself from other factors that contribute to answer variability, [] introduced the VQA-AnswerTherapy dataset. This dataset is the first to provide visual localization for all valid answers to each visual question. Its introduction not only offers more accurate annotations for answer localization in VQA tasks but also provides important insights into the discrepancies in how visual questions are interpreted by different annotators. Building upon the VQA-AnswerTherapy dataset, researchers further proposed an algorithmic challenge for answer localization, aimed at revisiting how diverse answer localizations should be handled during the development and evaluation of VQA models, rather than focusing solely on a single answer localization. This challenge emphasizes not only the incorporation of annotator variability into model training but also the advancement of ethical AI, ensuring that algorithms can more fairly and comprehensively understand and address the range of potential answers.
% 为了区分由视觉问题本身带来的固有挑战与其他导致答案多变因素，[]引入了 VQA-AnswerTherapy 数据集。该数据集是首个为每个视觉问题的所有有效答案提供视觉定位的数据。它的引入不仅为 VQA 任务中的答案定位提供了更准确的标注，同时也为不同标注者对视觉问题的理解差异提供了重要见解。在此基础上，研究人员进一步提出了答案定位的算法挑战[]，目标是推动VQA模型开发与评估时，重新审视如何应对多样化的答案定位问题，而不仅仅限于处理单一答案的定位。这个挑战不仅要求模型能够有效地处理注释者差异，并将其纳入训练过程，还特别关注如何通过这一过程推动伦理AI的发展，确保算法能够更加公正、全面地理解和处理多种潜在答案。
% 3. 问题陈述：现有模型难以处理多答案、多定位的情况.这段写得很好，但是需要和上一段有点衔接。
This phenomenon reveals two key challenges: 
(1) BLV users often cannot specify the exact image region of interest, leading to annotations over multiple, possibly disjoint areas; 
(2) assuming a single “ground-truth” answer in evaluation penalizes models for annotator disagreement. 
We formalize these as \textbf{visual uncertainty} caused by image artifacts like blur, poor lighting, and framing.  \textbf{Semantic ambiguity}, arising from open-ended, subjective questions. 
These factors result in high annotation variance: some answers align with different regions (\textit{multi-grounding}), others describe the same region differently (\textit{multi-expression}). 
Addressing this requires moving beyond standard VQA toward reasoning over visual-semantic consistency across diverse, valid answers in BLV-specific contexts.

To address these limitations, we propose \textbf{BLaVe-CoT}, a next-step VQA framework designed specifically for BLV users. Unlike conventional VQA systems that assume one clear answer, BLaVe-CoT embraces the reality of ambiguous, under-specified questions and aims to support scenarios where multiple visually grounded answers may be valid.  
It first uses a LoRA-tuned BLIP-2 model to propose a diverse set of plausible answers, then applies PolyFormer to extract the spatial evidence associated with each answer.  
A lightweight Chain-of-Thought (CoT) reasoning module then compares visual overlap and semantic similarity to decide whether the answers refer to the same or distinct regions—transforming VQA into a reasoning task about answer consistency rather than prediction alone.
By framing this as a visual–semantic consistency prediction task, our goal is not to replace existing VQA pipelines, but to extend them with essential reasoning capabilities needed for inclusive, real-world BLV-aware deployment.
\begin{figure*}
    \centering
    \includegraphics[width=0.75\linewidth]{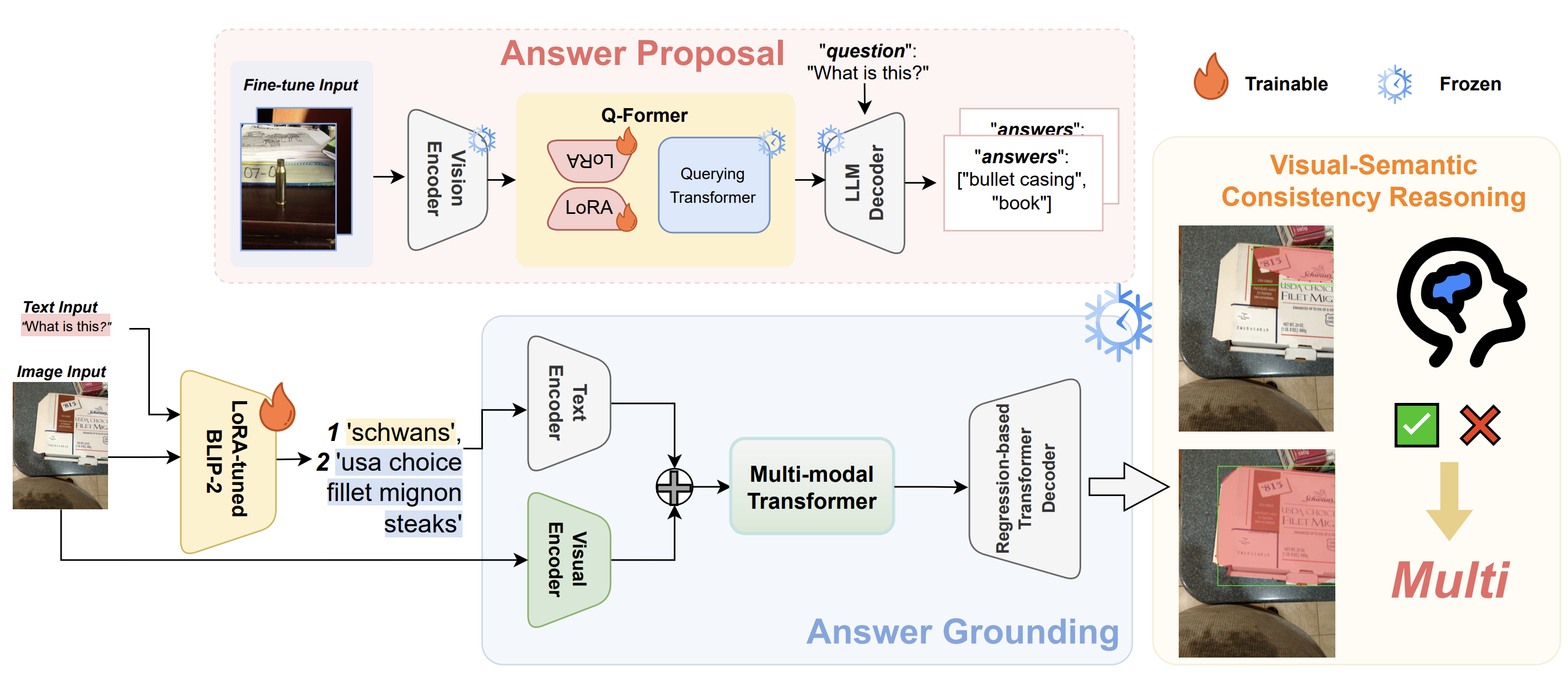}
    \caption{
Overview of the proposed BLaVe-CoT framework for ambiguity-aware Visual Question Answering. 
}

    \label{fig:enter-label}
\end{figure*}
Our contributions are summarized as follows:
\begin{itemize}
    \item We propose the Visual-Semantic Consistency Prediction task to explicitly address answer diversity in real-world VQA, particularly for blind and low vision users.
    \item We design BLaVe-CoT, a hybrid framework combining LoRA-tuned answer proposal, polygon-based answer grounding, and multimodal chain-of-thought reasoning.
    \item We achieve superior results on the VQA-AnswerTherapy benchmark, demonstrating robustness in BLV assistive scenarios.
\end{itemize}

\section{Related Work}

\subsection{BLV Visual Task}
The blind-vision task integrates computer vision and natural language processing technologies, aiming to enhance the environmental understanding of individuals with visual impairments \cite{wang2021survey}. In recent years, several solutions have been proposed for this task, such as object detection systems designed for blind and multi-object navigation technologies, which improve obstacle detection and navigation guidance \cite{meshram2019astute}, allowing visually impaired individuals to better comprehend their surroundings. Numerous researchers have developed human-driven visual explanation systems, which provide real-time object recognition and environmental analysis, assisting visually impaired users in obtaining contextual information about their environment \cite{varshney2025evaluating}. However, current computer vision research still pays limited attention to the BLV community. Our research offers significant benefits for visually impaired individuals by addressing semantic ambiguity and visual uncertainty in real-time question answering, thereby enhancing their ability to understand the environment.
% 盲人视觉任务结合了计算机视觉和自然语言处理技术，旨在帮助视觉障碍者提升对环境的理解能力。近年来，针对这一任务，例如，为盲人设计的物体检测系统和多物体导航技术增强了障碍物检测与导航引导功能[20,28,31]，帮助视力障碍者更好地理解周围环境。许多研究者开发了人力驱动的视觉解释系统，如Google Lookout和Envision，提供实时的物体识别和环境分析，协助视力障碍用户获取周围环境的上下文信息[5,1,2]。我们的研究可为视觉障碍者带来广泛益处，改善实时问答中语义模糊和视觉不确定性引起的理解误差，增强环境理解能力。

\subsection{BLV Visual Question Answering}
In recent years, numerous large-scale VQA datasets\cite{zou2020survey} have been proposed to support researchers in developing models capable of identifying and locating the visual evidence upon which humans rely when answering visual questions. However, previous VQA datasets are typically manually designed and often differ from real-world VQA scenarios. To better capture visual problems in real-world settings, the VizWiz dataset \cite{gurari2018vizwiz} was introduced as the first VQA dataset that reflects the genuine needs of blind users, with questions based on their visual perceptions. This dataset, created by blind photographers, contains unique challenges, including blurriness of the images, and approximately 28\% of the visual questions are considered difficult to answer by crowd-sourced workers due to the inherent characteristics of the data set. Building on VizWiz, the VQA-AnswerTherapy dataset \cite{chen2023vqa} takes a step further by providing multi-answer region annotations, enabling the study of ambiguous questions and diverse visual grounding in BLV VQA scenarios.

% 近年来，已有许多大规模的VQA数据集被提出，以支持研究人员开发能够识别和定位人类在回答视觉问题时所依赖的视觉证据的模型。然而，先前的VQA数据集通常是通过人工设计的，与实际应用中的VQA场景存在一定差距。为了更好地揭示真实场景中的视觉问题，VizWiz数据集应运而生，这是首个反映盲人用户真实需求并基于他们的视觉感知提出问题的VQA数据集。该数据集由盲人摄影师拍摄，由于其特殊性，数据集中的图像常常存在模糊问题，且约28%的视觉问题被众包工作者认为难以回答。
\subsection{Variability in Answers within VQA Datasets}
Answer diversity remains a long-standing challenge in VQA. Prior studies \cite{sterz2024dare} show that visual questions often have multiple valid answers due to subjective interpretation, contextual variation, and visual ambiguity. To address this, recent efforts \cite{khan2024consistency} have developed evaluation frameworks that acknowledge answer diversity and use trained models to predict potential labeling inconsistencies. From the perspective of blind VQA datasets, we introduce a visual-semantic consistency mechanism to predict answer diversity by examining whether distinct answers correspond to different visual regions.

\section{Method}
In this section, we detail the formulation and implementation of each component, highlighting how they work together to assess grounding consistency and improve model interpretability in ambiguous, multi-answer scenarios.
Figure~\ref{fig:enter-label} provides an overview of our proposed BLaVe-CoT pipeline. Each component is designed to explicitly handle the ambiguity and uncertainty common in BLV-oriented VQA settings.

\subsection{Problem Definition and Motivation}
Traditional Visual Question Answering models tend to learn a conditional distribution $P(A \mid I, Q)$, where $I$ denotes the image, $Q$ the natural language question, and $A$ the predicted answer. This formulation implicitly assumes a deterministic one-to-one mapping from the image-question pair $(I, Q)$ to a unique, unambiguous answer.

However, in real-world Blind or Low Vision scenarios, such an assumption often fails. Visual inputs may be degraded, ambiguous, or underspecified, leading to a range of plausible answers. These answers may describe different concepts or refer to different image regions, even when equally valid. Existing VQA systems lack mechanisms to reason over this diversity, undermining both reliability and interpretability.

We therefore reformulate the task as a visual grounding consistency prediction problem. Specifically, we aim to determine whether all valid answers for a given $(I, Q)$ pair refer to the same image region (\textit{single grounding}) or to distinct regions (\textit{multiple groundings}). Formally, the objective is to learn a function:
\begin{equation}
    f(I, Q) \rightarrow s \in [0, 1],
\end{equation}
where $s = 1$ indicates single grounding, and $s = 0$ indicates multiple groundings. This formulation shifts the focus from generating an answer to reasoning about answer alignment in the visual space.

\subsection{Answer Proposal via LoRA-tuned BLIP-2}
To generate candidate answers, we adopt BLIP-2 \cite{li2023blip}, a vision-language model known for robust answer generation. As illustrated in the top of Figure~\ref{fig:enter-label}, we fine-tune BLIP-2 using Low-Rank Adaptation (LoRA) \cite{hu2022lora}, applying updates only to the Query and Key projection matrices in the Q-Former, while freezing the vision encoder and OPT-2.7B decoder. LoRA approximates the weight update by $\Delta W = B A$, where $A \in \mathbb{R}^{r \times d}$ and $B \in \mathbb{R}^{d \times r}$ are low-rank matrices and $r \ll d$.

Given a pair $(I, Q)$, the tuned model generates top-$k$ candidate answers:
\begin{equation}
    A = \{a_1, a_2, \dots, a_k\} = M_{\text{VQA}}(I, Q).
\end{equation}
In our experiments, we set $k = 3$ to ensure diversity without introducing excessive noise. These answers serve as hypotheses for visual localization.

\begin{table*}[t]
\caption{\textbf{Performance on the VQA-AnswerTherapy benchmark.}  
All three metrics are evaluated on the \emph{“single”} class, so previous methods that always predict “single” can artificially boost Precision (and thus F$_1$).  
BLaVe-CoT shows the largest Recall gain, indicating it truly distinguishes multi-grounded questions rather than exploiting this bias.  
Bold = best; underline = second best.}

  \label{tab:main}
  \centering
  \setlength{\tabcolsep}{9pt}         % 控制列间距
  \renewcommand{\arraystretch}{1.25}  % 控制行高
  \begin{tabular}{lcccccc}
    \hline
    \textbf{Model} & \textbf{Reference} & \textbf{F$_1$} & \textbf{Precision} & \textbf{Recall} & \textbf{VQAv2 F$_1$} & \textbf{VizWiz F$_1$} \\
    \hline
    BLIP2-VizWiz    & ICML’23 \cite{li2023blip}          & 75.82 & 78.54 & 73.42 & 75.86 & 75.79 \\
    BLIP2           & ICML’23\,\cite{li2023blip} & 78.21 & 79.17 & 77.22 & \textbf{87.50} & 71.74 \\
    ViLT            & CVPR’24\,\cite{aboahvision}  & \underline{80.05} & \underline{80.40} & \underline{79.77} & 84.76 & \underline{77.20} \\
    \hline
    \textbf{BLaVe-CoT (ours)} & — & \textbf{82.63(\textcolor{red}{+2.58})} & \textbf{80.94(\textcolor{red}{+0.54})} & \textbf{84.33(\textcolor{red}{+4.56})} & \underline{87.20} & \textbf{79.83(\textcolor{red}{+2.63})} \\
    \hline
  \end{tabular}
\end{table*}

\subsection{Answer Grounding with PolyFormer}

To ground each candidate answer in the image, we employ PolyFormer \cite{liu2023polyformer}, a text-conditioned polygon segmentation model. As shown in Figure \ref{fig:enter-label}, given an image $I$ and a candidate answer $a_i \in A$, we construct a composite grounding query $t_i = Q + a_i$ by concatenating the original question $Q$ with the answer hypothesis. The image $I$ is encoded using a frozen visual encoder, and $t_i$ is simultaneously embedded via a frozen text encoder. These two feature streams are fused through a multi-modal transformer, whose output is fed into a regression-based decoder to predict a binary segmentation mask $m_i \in \{0,1\}^{H \times W}$:

\begin{equation}
    m_i = M_{\text{VG}}(I, t_i).
\end{equation}

The resulting mask $m_i$ highlights the spatial region corresponding to the candidate answer $a_i$. Collectively, the grounded masks $M = \{m_1, m_2, m_3\}$ serve as explicit visual evidence, enabling downstream reasoning over the spatial consistency of answer references.

\begin{figure}
    \centering
    \includegraphics[width=1.0\linewidth]{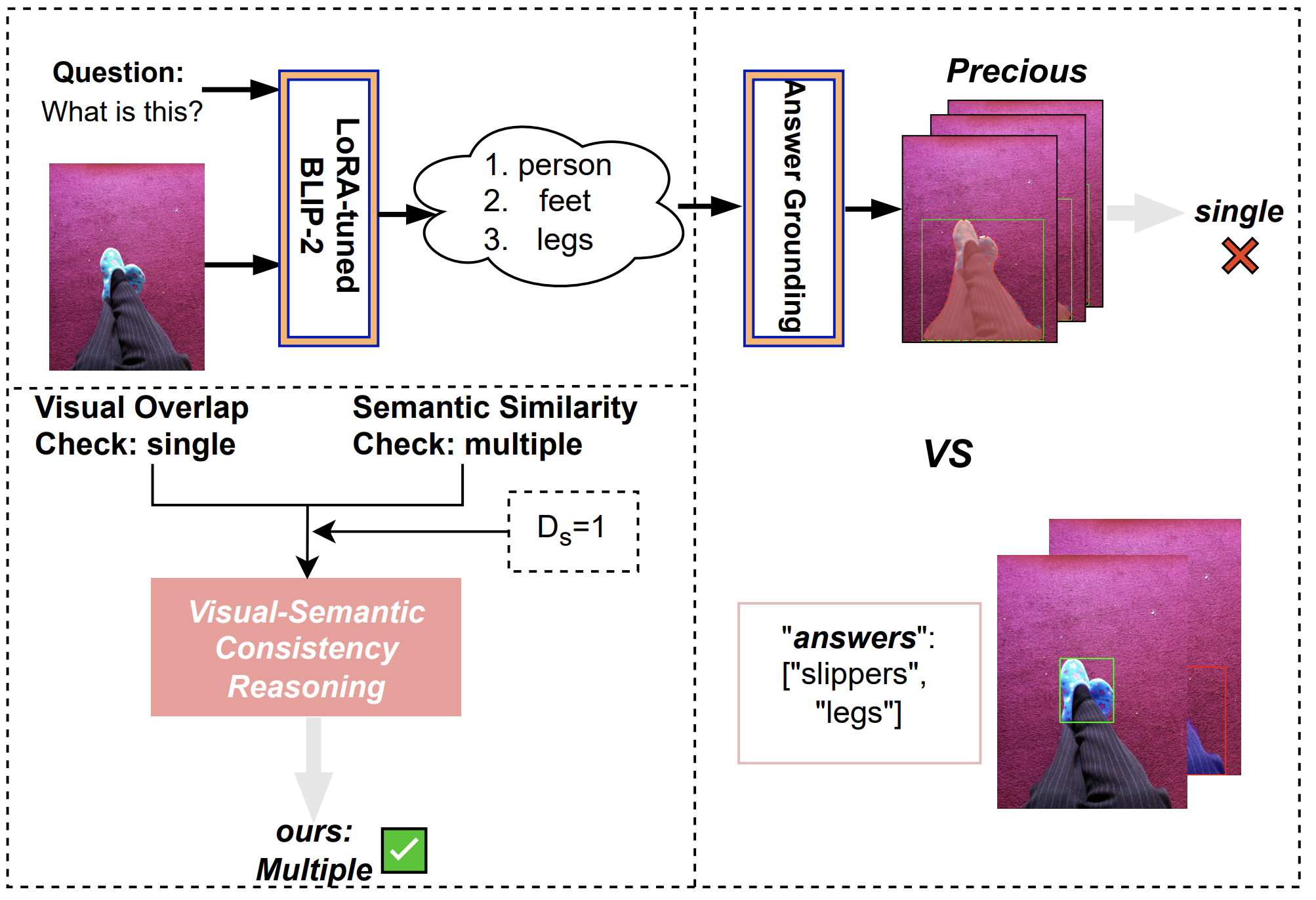}
\caption{
\textbf{The limitations of visual localization.} Due to the limitations of the VizWiz-AnswerTherapy training set, PolyFormer excessively relies on visual localization when processing VQA answers with different semantics, leading to incorrect answers. By incorporating CoT logic, which combines semantic information and visual masks for answer consistency evaluation, the model is able to more accurately recognize answer diversity.
}
    \label{fig:polyformer}
\end{figure}

\subsection{Visual-Semantic Consistency Reasoning}
Traditional single-stage classifiers are often hard to capture the disagreement patterns across diverse answer candidates in multi-annotator VQA settings. In contrast, we adapt a Chain-of-Thought reasoning structure that explicitly disentangles the two dominant sources of answer variation: visual grounding inconsistency and semantic ambiguity. This modular approach enables interpretable intermediate decisions, and provides robustness against edge cases—such as semantically divergent but visually similar answers, or numeric responses that lack strong language signal Figure~\ref{fig:polyformer}. By structuring the reasoning process into distinct substeps, BLaVe-CoT can better reflect the true alignment between multimodal evidence. We now determine whether the set $A$ is visually consistent. Our reasoning module $R$ performs multimodal comparison over $A$ and $M$.

\paragraph{Visual Consistency.}
We measure spatial overlap between masks using intersection-over-union (IoU):
\begin{equation}
    \text{IoU}(m_i, m_j) = \frac{|m_i \cap m_j|}{|m_i \cup m_j|}.
\end{equation}
We define a visual agreement indicator:
\begin{equation}
    C_V = \mathbb{I}\left( \min_{i \ne j} \text{IoU}(m_i, m_j) \ge \tau_{\text{iou}} \right),
\end{equation}
where $\tau_{\text{iou}}$ is a hyperparameter.

\paragraph{Semantic Consistency.}
We compute sentence embeddings $\mathbf{e}_i = E(a_i)$ using MiniLM \cite{wang2020minilm}, then derive pairwise cosine similarities:
\begin{equation}
    \text{sim}(a_i, a_j) = \frac{\mathbf{e}_i^\top \mathbf{e}_j}{\|\mathbf{e}_i\|_2 \cdot \|\mathbf{e}_j\|_2}.
\end{equation}
Semantic disagreement is indicated as:
\begin{equation}
    D_S = \mathbb{I}\left( \max_{i \ne j} \text{sim}(a_i, a_j) < \tau_{\text{sem}} \right).
\end{equation}

\paragraph{Decision Logic.}
We define the final consistency score $s$ as:
\begin{equation}
    s =
    \begin{cases}
        C_V, & \text{if all } a_i \text{ are numeric}, \\\\
        0, & \text{if } D_S = 1, \\\\
        C_V, & \text{otherwise}.
    \end{cases}
\end{equation}
This prioritizes semantic distinction in general cases and defers to visual evidence when dealing with numeric answers (e.g., \texttt{"two"}, \texttt{"3"}).

\paragraph{End-to-End Pipeline.}
The full pipeline is summarized as:
\begin{equation}
    (I, Q) \xrightarrow{M_{\text{VQA}}} A \xrightarrow{M_{\text{VG}}} M \xrightarrow{R} s.
\end{equation}

\begin{algorithm}[t]

\caption{CoT-Based Visual-Semantic Consistency Reasoning}
\label{alg:cot}
\KwIn{Candidate answers $A = \{a_1, a_2, a_3\}$, masks $M = \{m_1, m_2, m_3\}$}
\KwOut{Consistency score $s \in \{0, 1\}$}

\BlankLine
\tcp{Step 1: Visual Overlap Check}
$C_V \gets \mathbb{I} \left( \min_{i \ne j} \text{IoU}(m_i, m_j) \ge \tau_{\text{iou}} \right)$

\tcp{Step 2: Semantic Similarity Check}
Compute sentence embeddings: $\mathbf{e}_i = E(a_i)$ for each $a_i \in A$ \\
$D_S \gets \mathbb{I} \left( \max_{i \ne j} \text{sim}(\mathbf{e}_i, \mathbf{e}_j) < \tau_{\text{sem}} \right)$

\tcp{Step 3: Special Case for Numeric Answers}
\If{all $a_i$ are numeric}{
    $s \gets C_V$ \tcp*{Trust visual alignment for numbers}
}
\ElseIf{$D_S = 1$}{
    $s \gets 0$ \tcp*{Semantic disagreement indicates multiple grounding}
}
\Else{
    $s \gets C_V$ \tcp*{Default to visual consistency}
}

\Return $s$
\end{algorithm}

\begin{figure}
    \centering
    \includegraphics[width=0.80\linewidth]{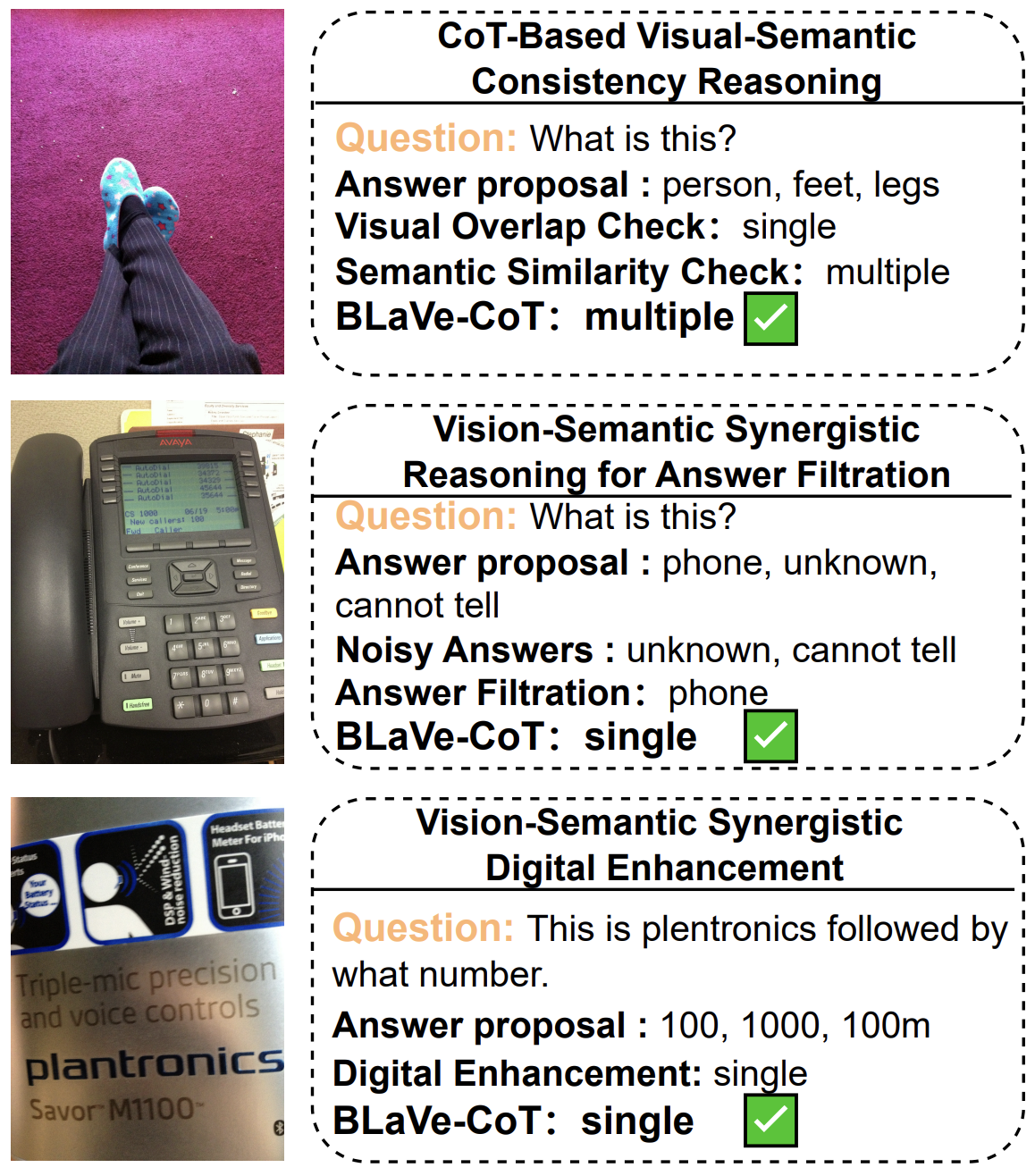}
\caption{
The CoT-Based Visual-Semantic consistency reasoning include: (1) \textbf{Visual-Semantic inconsistency:} When the visual-semantic prediction results are inconsistent and the semantic similarity falls below a threshold, the semantic predicted answer is selected; (2) \textbf{Answer Filtration: }When the BLIP model generates meaningless answers, COT filters them and retains only valid answers; (3) \textbf{Digital Enhancement: }Special handling of numbers by using mask predictions as the final answer, ensuring that when BLIP predictions are incorrect, the image mask helps to improve prediction accuracy.
}
    \label{fig:cot-based}
\end{figure}

The overall CoT-based reasoning logic is summarized in Algorithm~\ref{alg:cot}, providing a clear step-by-step view of how visual and semantic signals are fused to reach the final grounding decision.
As shown in Figure~\ref{fig:cot-based}, this modular framework facilitates fine-grained reasoning over answer diversity, improving VQA transparency in BLV contexts.

\section{Experiments}

\subsection{Experimental Setup}

\subsubsection{Dataset}

We evaluate our proposed BLaVe-CoT framework on the VQA-AnswerTherapy benchmark and report comprehensive results in Table~\ref{tab:main}. Our method is compared against baselines including BLIP2, BLIP2 fine-tuned on VizWiz (BLIP2-VizWiz), and ViLT fine-tuned on VizWiz (ViLT-VizWiz), across multiple metrics such as overall F$_1$, Precision, Recall, as well as subset-level performance on VQAv2 and VizWiz.

\subsubsection{Evaluation Metrics}
To be consistent with prior work on VQA-AnswerTherapy \cite{chen2023vqa,aboahvision}, we evaluate whether the model correctly judges a question as having a \emph{single} or \emph{multiple} valid groundings with the standard trio of Precision, Recall and F$_1$:
\begin{itemize}
    \item \textbf{Precision} ($P$): The proportion of correctly predicted “single” instances among all instances predicted as “single”.
    \begin{equation}
        P = \frac{\text{\# correctly predicted ``single''}}{\text{\# predicted ``single''}}
    \end{equation}
    
    \item \textbf{Recall} ($R$): The proportion of correctly predicted “single” instances among all ground-truth “single” instances.
    \begin{equation}
        R = \frac{\text{\# correctly predicted ``single''}}{\text{\# actual ``single''}}
    \end{equation}
    
    \item \textbf{F$_1$ Score}: The harmonic mean of Precision and Recall, which captures the balance between these two aspects.
    \begin{equation}
        \text{F$_1$} = \frac{2 \cdot P \cdot R}{P + R}
    \end{equation}
\end{itemize}

These metrics provide a comprehensive view of the model’s ability to make correct binary judgments about single-answer visual grounding scenarios.

\subsection{Results}
As shown in Table~\ref{tab:main}, BLaVe-CoT achieves better performance compared to competitors in multiple evaluation metrics and datasets.

\textbf{Overall performance.} BLaVe-CoT achieves the best overall F$_1$ score of 82.63, outperforming the strongest baseline (ViLT, 80.05) by a margin of +2.58. This consistent improvement is further reflected in both overall Recall and Precision. Notably, our model attains an overall Recall of 84.33, exceeding the ViLT by +4.56, indicating superior ability to retrieve semantically valid answers. The Precision also improves to 80.94. 

\textbf{Subset-level analysis.}
On the \textit{VQAv2} split, BLaVe-CoT attains an F$_1$ of 87.20, virtually matching BLIP-2’s 87.50 and exceeding ViLT by +2.44. BLIP-2’s slight edge is expected, as it was pretrained on large vision-language corpora whose distribution closely mirrors VQAv2, giving it a dataset-specific bias rather than broader reasoning strength.
The difference emerges on the more demanding \textit{VizWiz} split—low-quality photos taken by BLV users—where BLaVe-CoT scores 79.83, surpassing ViLT (77.20) and BLIP-2 (71.74) by up to +2.63. This confirms our model’s superior adaptability and grounding robustness under real-world, high-variance conditions.

\subsection{Ablation Study}

To evaluate the contribution of each core component in the proposed BLaVe-CoT framework, we conduct an ablation study focusing on two critical modules: (1) \textit{Answer Proposal via LoRA-tuned BLIP-2}, and (2) \textit{Visual-Semantic Consistency Reasoning}. Four model variants are compared, with results summarized in Table~\ref{tab:ablation-consistent} and visualized in Figure~\ref{fig:ablation}.
\begin{table}[t]
\caption{\textbf{Ablation study on answer proposal and reasoning modules.} \ding{51} denotes that the component is enabled. BLaVe-CoT includes both fine-tuned answer proposal and CoT-based reasoning.}
\label{tab:ablation-consistent}
\centering
\setlength{\tabcolsep}{4.5pt}
\renewcommand{\arraystretch}{1.2}
\begin{tabular}{l c c | c c c}
\hline
\toprule
\textbf{Method} & \textbf{LoRA} & \makecell{\textbf{CoT} \\ \textbf{Reasoning $R$}} & \textbf{F$_1$} & \textbf{Precision} & \textbf{Recall} \\
\midrule
\hline
BLIP2 (Frozen)         & \ding{55} & \ding{55} & 78.21 & 79.17 & 77.22 \\
LoRA-tuned     & \ding{51} & \ding{55} & 79.75 & 79.60 & 79.75 \\
CoT Reasoning          & \ding{55} & \ding{51} & 80.75 & 79.22 & 82.28 \\
\textbf{BLaVe-CoT}     & \ding{51} & \ding{51} & \textbf{82.63} & \textbf{80.94} & \textbf{84.33} \\
\bottomrule
\hline
\end{tabular}
\end{table}

The plain BLIP-2 backbone achieves an F$_1$ of 78.21.
LoRA fine-tuning alone lifts this to 79.75, showing that domain adaptation helps the model propose more appropriate answers for BLV data.
Replacing LoRA with our visual–semantic reasoning module raises F$_1$ further to 80.75 (recall 82.28), confirming that structured consistency checks reduce answer noise.
When the two components are used together, performance peaks at 82.63, demonstrating that LoRA-based answer generation and consistency reasoning are complementary.
\subsection{Training Stability Across Epochs}

\begin{figure}[t]
    \centering
    \begin{subfigure}[b]{0.45\linewidth}
        \centering
        \includegraphics[width=\linewidth]{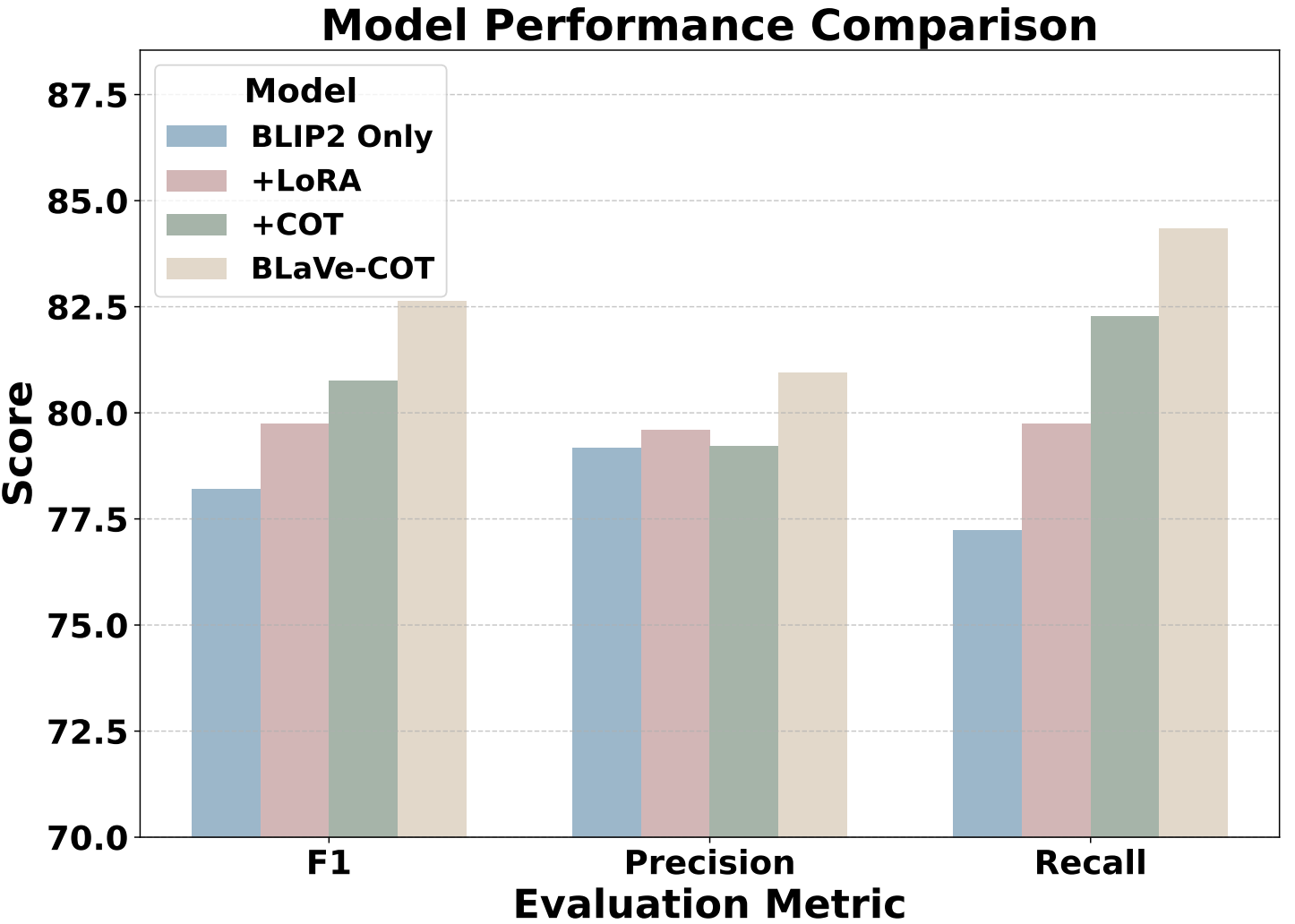}
        \caption{Component-wise Comparison}
        \label{fig:bar_ablation}
    \end{subfigure}
    \hfill
    \begin{subfigure}[b]{0.45\linewidth}
        \centering
        \includegraphics[width=\linewidth]{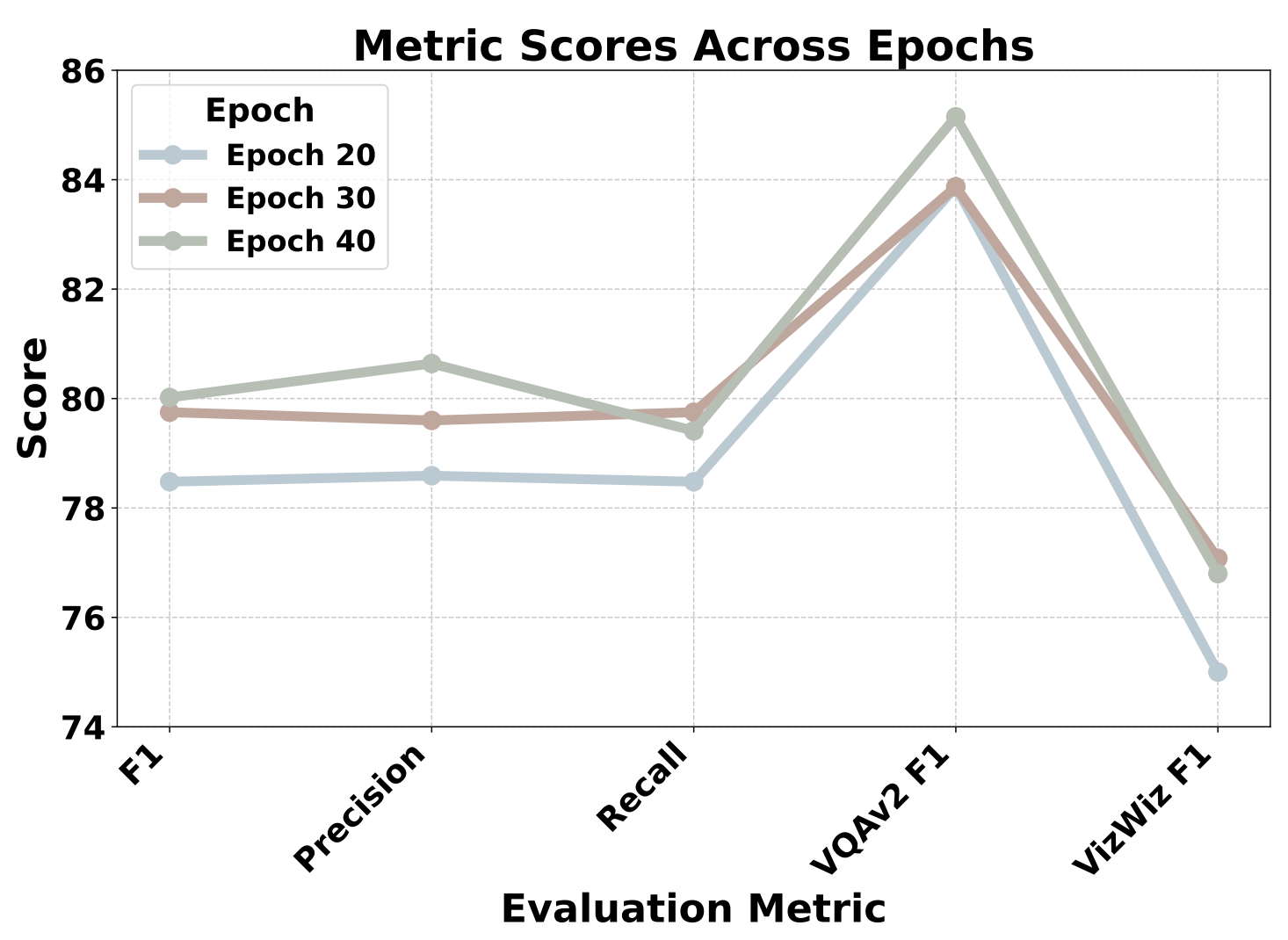}
        \caption{Performance Across Epochs}
        \label{fig:line_epochs}
    \end{subfigure}
    \caption{\textbf{Ablation Results.} (a) Effect of LoRA and reasoning modules. (b) Performance trend across training epochs.}
    \label{fig:ablation}
\end{figure}

As shown in Figure~\ref{fig:line_epochs}, performance improves steadily across training epochs. We select Epoch 40 as the final checkpoint, as it consistently delivers the best results across metrics.

\section{CONCLUSIONS}
This paper introduces BLaVe-CoT, a consistency-aware Visual Question Answering framework specifically tailored to the unique challenges faced by blind and low-vision users. By explicitly addressing visual uncertainty and semantic ambiguity prevalent in real-world BLV contexts, BLaVe-CoT effectively moves beyond traditional single-answer paradigms. Key components include LoRA-tuned BLIP-2 for diverse answer generation, PolyFormer for precise answer grounding, and a structured Chain-of-Thought reasoning module for visual-semantic consistency assessment. Experimental results on the VQA-AnswerTherapy dataset confirm that our method significantly outperforms state-of-the-art models, demonstrating its ability to robustly handle multi-answer, multi-region scenarios.

Future work will focus on the real-world challenge that BLV data are scarce and highly variable. We plan to: introduce multi-granularity confidence alignment to improve cross-domain robustness \cite{chen2025refining}. At the same time, few-shot learning techniques will help us cut down the need for large-scale manual labels \cite{ruan2024advances, zanxi2024}. And draw on open-set category discovery and knowledge reasoning to handle previously unseen concepts \cite{pu2023dynamic}.

All these directions aim at one goal: keeping BLaVe-CoT reliable and continually evolving in data-sparse, ever-changing BLV settings.

\addtolength{\textheight}{-12cm}   % This command serves to balance the column lengths

%%%%%%%%%%%%%%%%%%%%%%%%%%%%%%%%%%%%%%%%%%%%%%%%%%%%%%%%%%%%%%%%%%%%%%%%%%%%%%%%

%%%%%%%%%%%%%%%%%%%%%%%%%%%%%%%%%%%%%%%%%%%%%%%%%%%%%%%%%%%%%%%%%%%%%%%%%%%%%%%%



% \section*{ACKNOWLEDGMENT}

% The preferred spelling of the word ÒacknowledgmentÓ in America is without an ÒeÓ after the ÒgÓ. Avoid the stilted expression, ÒOne of us (R. B. G.) thanks . . .Ó  Instead, try ÒR. B. G. thanksÓ. Put sponsor acknowledgments in the unnumbered footnote on the first page.

%%%%%%%%%%%%%%%%%%%%%%%%%%%%%%%%%%%%%%%%%%%%%%%%%%%%%%%%%%%%%%%%%%%%%%%%%%%%%%%%

\normalem
\bibliographystyle{IEEEtran}   % IEEE 推荐的 BibTeX 样式
\bibliography{IEEEexample}            % ref.bib 里存放所有文献

\end{document}